\pdfoutput=1

\documentclass[11pt]{article}

\usepackage{acl}

\usepackage{times}
\usepackage{graphicx}
\usepackage{latexsym}
\usepackage{natbib}
\usepackage{amsmath}
\usepackage{enumerate}
\usepackage{amssymb}
\usepackage{multirow}
\usepackage{multicol}
\usepackage{enumitem}
\usepackage{booktabs}
\usepackage{enumitem}
\usepackage{balance}
\usepackage{natbib}
\usepackage[T1]{fontenc}

\usepackage[utf8]{inputenc}

\usepackage{microtype}

\title{KCD: Knowledge Walks and Textual Cues Enhanced \\ Political Perspective Detection in News Media}

\author{Wenqian Zhang$^\spadesuit$\thanks{\ \ These authors contributed equally to this work.} \: \: 
Shangbin Feng$^{\spadesuit}$\footnotemark[1] \: \: 
Zilong Chen$^{\diamondsuit}$\footnotemark[1] \: \: \\ \bf Zhenyu Lei$^{\diamondsuit}$ \: \: 
Jundong Li$^{\heartsuit}$ \: \:  Minnan Luo$^{\spadesuit}$\thanks{\ \ Corresponding author.}
\\
School of Computer Science and Technology, Xi’an Jiaotong University$^{\spadesuit}$ \\ School of Science, Xi’an Jiaotong University$^{\diamondsuit}$, University of Virginia$^{\heartsuit}$ \\ \texttt{\{2194510944,wind\_binteng,luoyangczl,Fischer\}@stu.xjtu.edu.cn}\\ \texttt{jundong@virginia.edu} \: \:  \texttt{minnluo@xjtu.edu.cn}}

\begin{document}
\maketitle

\begin{abstract}
    Political perspective detection has become an increasingly important task that can help combat echo chambers and political polarization. Previous approaches generally focus on leveraging textual content to identify stances, while they fail to reason with background knowledge or leverage the rich semantic and syntactic textual labels in news articles. In light of these limitations, we propose KCD, a political perspective detection approach to enable multi-hop knowledge reasoning and incorporate textual cues as paragraph-level labels. Specifically, we firstly generate random walks on external knowledge graphs and infuse them with news text representations. We then construct a heterogeneous information network to jointly model news content as well as semantic, syntactic and entity cues in news articles. Finally, we adopt relational graph neural networks for graph-level representation learning and conduct political perspective detection. Extensive experiments demonstrate that our approach outperforms state-of-the-art methods on two benchmark datasets. We further examine the effect of knowledge walks and textual cues and how they contribute to our approach's data efficiency.
\end{abstract}
\section{Introduction}
Political perspective detection aims to identify ideological stances of textual data such as social media posts and news articles. Previous approaches generally leverage the textual content of news articles with various text modeling techniques to identify political stances. 
Those works \citep{CNNglove, li2019encoding, li2021mean, feng2021knowledge} leveraged diversified text models, such as recurrent neural networks \citep{HLSTM}, word embedding techniques \citep{pennington2014glove, ELMo}, convolutional neural networks \citep{CNNglove}, and pre-trained language models \citep{devlin2018bert, liu2019roberta}, to encode news paragraphs and classify them into different perspective labels.
Later approaches incorporate information sources beyond text to facilitate argument mining and boost task performance. News discussion on social networks \citep{li2019encoding}, social and linguistic information about news articles \citep{li2021mean}, media sources and information \citep{baly-etal-2020-detect} as well as external knowledge from knowledge graphs \citep{feng2021knowledge} are introduced in the task of political perspective detection and achieve better performance.

\begin{figure}[t]
    \centering
    \includegraphics[width=1\linewidth]{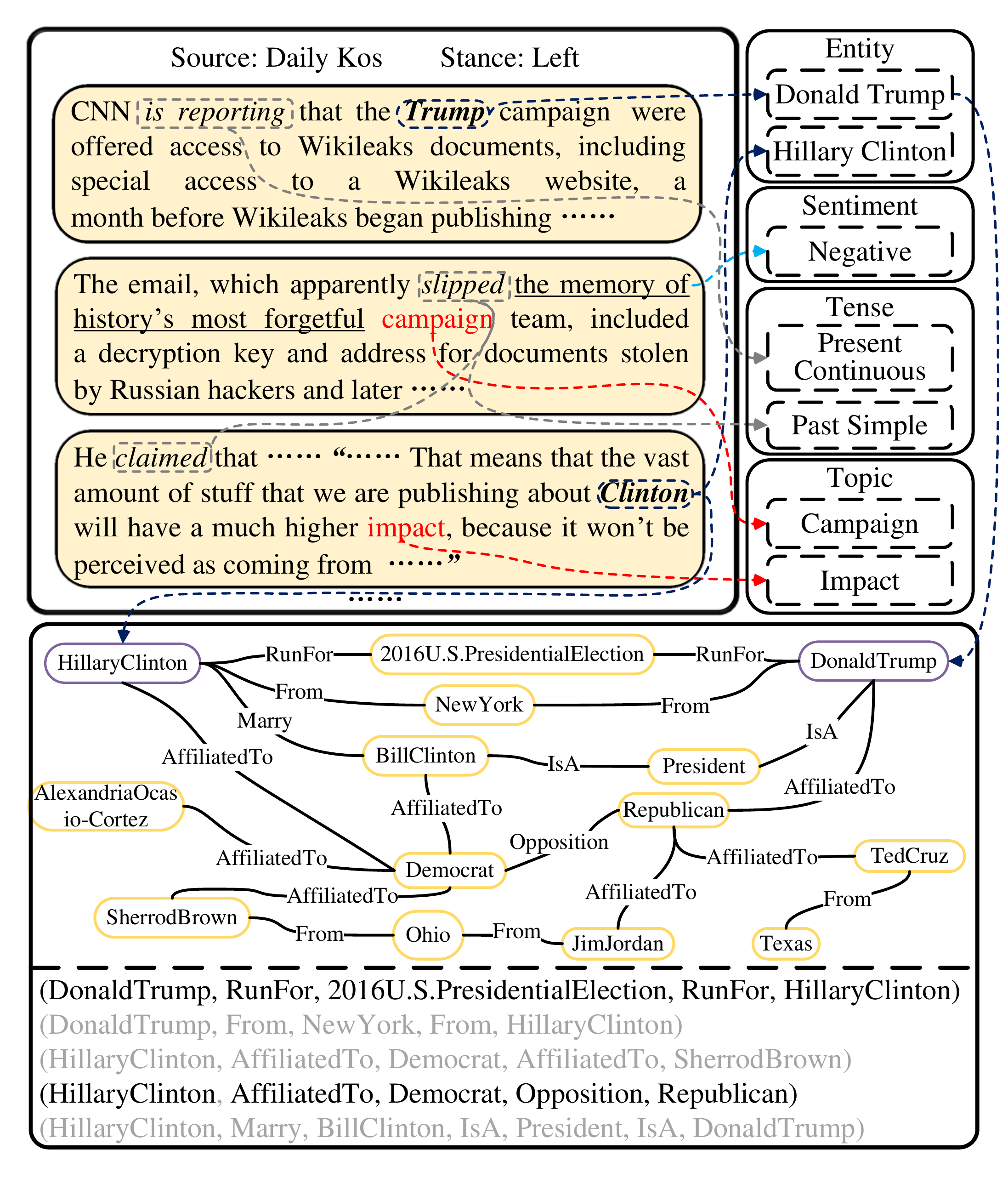}
    \caption{Multi-hop knowledge reasoning and implicit textual indicators that facilitate perspective detection.}
    \label{fig:teaser}
\end{figure}

Although these methods attempted to leverage more than news content, they fail to present a framework capable of reasoning with background knowledge and leveraging implicit semantic and syntactic indicators such as sentiment and tense of news articles. For example, Figure \ref{fig:teaser} presents a typical news article from Daily Kos\footnote{https://www.dailykos.com/}. This article discusses remarks from the Trump campaign team about Wikileaks and its effect on Hillary Clinton's bid for president. Individuals often rely on the multi-hop reasoning that Clinton and Trump are from opposite political parties and run against each other to inform their perspective analysis process. Besides, the negative sentiment expressed in satiric tones and the quotation of Trump campaign staff also give away the author's denial and left-leaning perspective. That being said, knowledge reasoning and implicit textual indicators are essential in the news bias detection process.

\begin{figure*}[t]
    \centering
    \includegraphics[width=\linewidth]{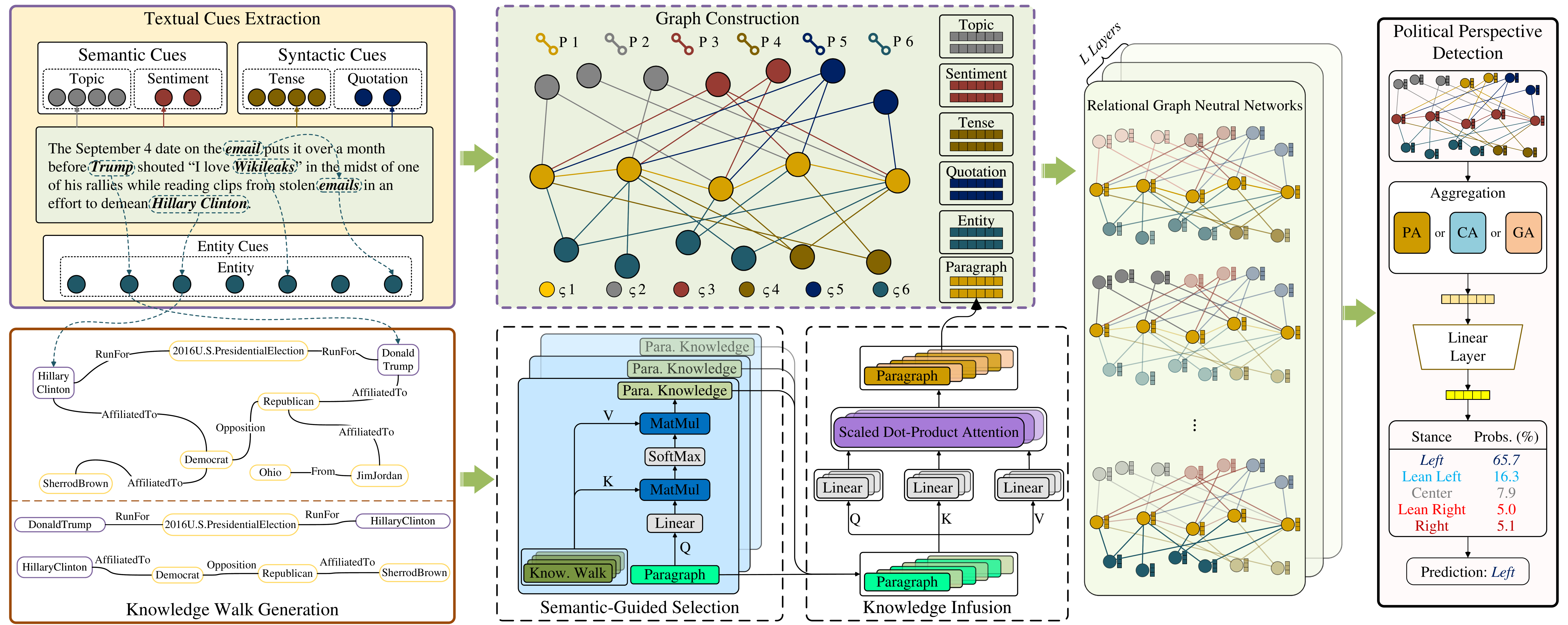}
    \caption{Overview of our proposed framework KCD.}
    \label{fig:overview}
\end{figure*}

In light of these limitations, we propose a political perspective detection framework \textbf{KCD} (\textbf{K}nowledge Walks and Textual \textbf{C}ues Enhanced Political Perspective \textbf{D}etection). Specifically, KCD generates multi-hop knowledge walks, aggregates them based on semantic relevance and incorporates them in textual representations with multi-head attention. KCD then constructs a heterogeneous information network to jointly model knowledge-enriched news content and diversified textual cues as paragraph-level labels. Finally, KCD learns graph representations with relational graph neural networks and conduct perspective detection with different 
aggregation strategies. Our main contributions are summarized as follows:

\begin{itemize}[leftmargin=*]
    \item We propose knowledge walks, a strategy to incorporate multi-hop knowledge reasoning in textual representations for knowledge-aware political perspective detection.
    \item We propose to construct a heterogeneous information network to represent news articles, which jointly models knowledge-enriched news content and implicit textual cues in news articles.
    \item Extensive experiments demonstrate that our approach consistently outperforms state-of-the-art methods on two widely adopted benchmarks. Further analysis bears out the necessity of knowledge walks and textual cues in our approach.
\end{itemize}

\section{Related Work}
\subsection{Political Perspective Detection}
Political perspective detection aims to identify the ideological stances of news articles, which is widely studied to help strengthen the online information landscape \citep{li2019encoding} and mitigate ideological echo chambers \citep{li2021mean, feng2021knowledge}. Early approaches leverage text analysis techniques for bias detection, such as sentiment analysis \citep{CIKMstance17, CIKMstance32}, bias feature extraction \citep{MEANbiasfeature}, word embeddings \citep{CNNglove, li2019encoding}, and different neural network architectures \citep{augenstein-etal-2016-stance, CIKMstance9, CIKMstance33, HLSTM, CNNglove, Feng2021LegislatorRL, li2021mean, feng2021knowledge}. In addition to textual content of news articles, social media users also become the focus of perspective detection research \citep{bel2021negation}. User interactions \citep{CIKMstance23}, user clustering \citep{CIKMstance7}, and label propagation \citep{CIKMstance29} are leveraged to identify the ideological preferences on social media. Fusing both news text and social network analysis directions, \citet{li2019encoding} propose to enrich news text with the content and structure of social media discussions about these news articles. Recent state-of-the-art approaches chart a new path by incorporating social and political external knowledge into stance detection.
\citet{baly-etal-2020-detect} propose adversarial media adaptation and leverage source background knowledge for political perspective detection.
\citet{li2021mean} combine language encoders with pre-training tasks of social and linguistic information.
\citet{feng2021knowledge} propose to construct and leverage political knowledge graphs as domain-specific external knowledge. In this paper, we build on these works to examine and explore the effect of multi-hop knowledge reasoning and diversified textual cues in the task of political perspective detection.

\subsection{Knowledge Graph in NLP}
Knowledge graphs (KGs) are effective representations of real-world entities, relations, and knowledge. Generic \citep{fellbaum2010wordnet, tanon2020yago, bollacker2008freebase, speer2017conceptnet} and domain-specific KGs \citep{feng2021knowledge, chang-etal-2020-benchmark} are widely adopted in NLP tasks as external knowledge sources. These approaches could mainly be categorized into feature extraction, language model and graph-based methods. For feature extraction approaches, KG embedding technique TransE \citep{TransE} is leveraged to learn features for knowledge injecton \citep{ostendorff2019enriching, hu2021knowledgefakenews}. For language model approaches, the adapter architecture is leveraged to fine-tune on KG-related tasks \citep{majewska2020verb, meng2021mixture, wei2021knowledge}. In addition, \citet{wang2021kepler} propose a unified model to combine knowledge embedding with language representation pre-training. For graph-based approaches, KG entities and relations are injected into graphs and heterogeneous information networks \citep{hu2021knowledgefakenews, feng2021knowledge, lu2021kelm}. Graph neural networks are then adopted to learn knowledge-aware text representations. In this paper, we propose knowledge walk, a novel strategy to infuse multi-hop knowledge reasoning into language representations and apply them in political perspective detection.

\section{Methodology}
Figure \ref{fig:overview} presents an overview of our proposed political perspective detection framework \textbf{KCD} (\textbf{K}nowledge Walks and Textual \textbf{C}ues Enhanced Political Perspective \textbf{D}etection). We firstly generate knowledge walks on the external knowledge graph.
These knowledge walks are then selected based on semantic relevance and injected into textual representations with multi-head attention. We then construct a heterogeneous information network to jointly model knowledge-enriched news content and diversified textual cues as paragraph-level labels and supernodes. Finally, we adopt relational graph neural networks and different aggregation strategies to learn graph-level representation and conduct political perspective detection.

\subsection{Knowledge Walks and Infusion}
\label{subsec:knowledgewalk}
We firstly propose the novel strategy of knowledge walks and combine them with textual representations to enable multi-hop knowledge reasoning. We partition an $n$-paragraph news document into different paragraphs and denote them as $\mathcal{S}=\{s_1,...,s_n\}$. We encode each paragraph by averaging the the embeddings of words from pre-trained RoBERTa \citep{liu2019roberta}:
\begin{equation}
\label{equ:first_RoBERTa}
{v}_{i}^s=RoBERTa(s_{i}), \ \ \ 1 \leq i \leq n
\end{equation}

We use a political knowledge graph\footnote{https://github.com/BunsenFeng/news\_stance\_detection} as external knowledge for perspective detection. Let the $i$-th triple in the knowledge graph be ${(e_{ih}, r_i, e_{it})}$, where $e_{ih}$ and $e_{it}$ denote the head and tail entity and $r_i$ represents the relation of the $i$-th triple.

\subsubsection{Knowledge Walk Generation}
We firstly use TagMe \citep{tagme} to identify mentioned KG entities in each paragraph $s_i$. For each mentioned entity, we use it as the starting point $e_{(0)}$ in a $K$-hop knowledge walk:

\begin{equation}
kw_i = \{e_{(0)},r_{0,1},e_{(1)}, ... , r_{K-1,K}, e_{(K)}\}
\end{equation}
where $e_{(i-1)}$ and $r_{i-1,i}$ denote the $i$-th triple's head entity and relation. Specifically, a knowledge walk is generated by adopting biased random walk of length $K$ starting from $e_{(0)}$. The conditional probability of arriving at $e_{(i)}$ from $e_{(i -1)}$ through $r_{i-1,i}$ is formulated as
\begin{equation}
    P(e_{(i)}\vert e_{(i-1)}, r_{i-1,i}) = \frac{exp(p(r_{i-1,i}))}{\sum_{j=1}^{\vert N_r(i-1)\vert} exp(p(r_j))}
\end{equation}
where $N_r(i-1)$ denotes the neighboring relations of $e_{(i-1)}$, $p(r)$ is the importance score of KG relation $r$, which could be tuned by domain experts for human-in-the-loop knowledge walk generation. In this way, we generate multiple knowledge walks for each paragraph based on its mentioned entities, which models the multi-hop reasoning process with external knowledge.

\subsubsection{Semantic-Guided Selection}
After obtaining multiple knowledge walks for a single news paragraph, we propose a selection and aggregation process guided by textual content to differentiate essential knowledge walks from the irrelevant ones. We firstly transform each knowledge walk $kw_i$ into a sentence $t_i$ by concatenating the textual description of entities and relations. We then encode the knowledge walk sentence $t_i$ with pre-trained RoBERTa \citep{liu2019roberta}:

\begin{equation}
{v}_{i}^k=RoBERTa(t_{i})
\end{equation}

Suppose a total of $m$ knowledge walks $\{kw_{i,j}\}_{j=1}^m$ are generated for paragraph $s_i$, we then aggregate their knowledge walk sentence embeddings $\{v_{i,j}^k\}_{j=1}^m$ as follows:
\begin{equation}
\label{equ:first_agg_start}
 v_i^p=\sum_{j=1}^{m}\frac{exp(\alpha \cdot v_{i,j}^k)}{\sum_{q=1}^{m}exp(\alpha \cdot v_{i,q}^k)} v_{i,j}^k
\end{equation}
where $\alpha$ denotes the learnable attention vector guided by paragraph semantics:
\begin{equation}
\label{equ:first_agg_end}
\alpha = \phi(W_{a}v_i^s+b_{a})    
\end{equation}
where $W_a$ and $b_a$ are learnable parameters of the attention module and we use Leaky-ReLU for $\phi$. $v_i^s$ is the sentence embedding from equation \ref{equ:first_RoBERTa}. In this way, we aggregate $m$ knowledge walks based on semantic relevance to the paragraph to filter and retain important knowledge reasoning paths.

\subsubsection{Knowledge Infusion}
After representing multi-hop knowledge reasoning for paragraph $s_i$ with $v_i^p$, we conduct document-wise multi-head self-attention to infuse knowledge walks into textual representations $v_i^s$. We concatenate knowledge walk and text representations:

\begin{equation}
\label{equ:second_agg_start}
    T = concat([v_1^s, v_1^p, ..., v_n^s, v_n^p])
\end{equation}
where $T$ is the input for multi-head self-attention:

\begin{equation}
\label{equ:second_agg_end}
\tilde{T}=MultiHead(Q, K, V)
\end{equation}
where $Q=K=V=T$ and the output $\tilde{T} = concat([\tilde{v_1^s}, \tilde{v_1^p}, ..., \tilde{v_n^s}, \tilde{v_n^p}])$. In this way, we obtain language representations of news paragraphs $\{\tilde{v_i^s}\}_{i=1}^n$, which jointly models textual content and related multi-hop knowledge reasoning paths.

\subsection{Textual Cues and Graph Construction}
We construct a heterogeneous information network (HIN) as in Figure \ref{fig:overview} ``Graph Construction`` to jointly represent knowledge-enriched news content and diversified textual cues in news articles. Specifically, we use paragraph nodes to represent the news content and connect them with different paragraph-level labels with heterogeneous edges. Firstly, for paragraph nodes:

\noindent \underline{\textit{$\mathcal{V}1$ and $\mathcal{R}1$: Paragraph Nodes}} We use one node in $\mathcal{V}1$ to represent each paragraph in the news article to partition the entire document and allow fine-grained analysis. We adopt the knowledge-enriched representations $\{\tilde{v_i^s}\}_{i=1}^n$ in Section \ref{subsec:knowledgewalk} as initial node features for $\mathcal{V}1$. We then use relation $\mathcal{R}1$ to connect adjacent paragraphs to preserve the original flow of the news article.

\subsubsection{Semantic Cues}
We further analyze the topic and sentiment of news paragraphs, extract paragraph-level labels and inject them into our news HIN structure.

\noindent \underline{\textit{$\mathcal{V}2$ and $\mathcal{R}2$: Topic Cues}} The topics and frequent topic switching in news articles often give away the stance and argument of authors. We train LDA to extract the topics in each political perspective detection corpus and use one node to represent each topic. We then encode the topic text with pre-trained RoBERTa as node attributes. We then use $\mathcal{R}2$ to connect each paragraph node in $\mathcal{V}1$ with its affiliated topic node in $\mathcal{V}2$ with the help of BertTopic \citep{grootendorst2020bertopic}.

\noindent \underline{\textit{$\mathcal{V}3$ and $\mathcal{R}3$: Sentiment Cues}} The sentiment of news articles signal the authors' approval or denial, which helps identify their stances towards individuals and issues. We use two nodes to represent positive and negative sentiment and we make their node attributes learnable. We then conduct sentiment analysis \citep{wolf-etal-2020-transformers} to identify paragraph sentiment and use $\mathcal{R}3$ to connect $\mathcal{V}1$ with their corresponding sentiment nodes in $\mathcal{V}3$.

\subsubsection{Syntactic Cues}

Apart from semantic cues, syntactic information in news articles also contribute to the perspective analysis process \citep{dutta2022semi}. In light of this, we analyze the tense of news paragraphs and whether it contains direct quotation and use them as paragraph-level labels in our constructed HIN.

\noindent \underline{\textit{$\mathcal{V}4$ and $\mathcal{R}4$: Tense Cues}} The tense of news paragraphs helps separate facts from opinions. For example, simple past tense often indicates factual statements while simple future tense suggests opinions and projections that might not be factual. We use 17 nodes in $\mathcal{V}4$ to represent 17 possible tenses in our constructed news HIN. We use NLTK \citep{nltk} to extract paragraph tenses and use $\mathcal{R}4$ to connect paragraph nodes in $\mathcal{V}1$ with $\mathcal{V}4$.

\noindent \underline{\textit{$\mathcal{V}5$ and $\mathcal{R}5$: Quotation Cues}} It is common for authors to directly quote others' words in news articles, which helps to identify the basis of the author's argument. We use two nodes to differentiate between whether a news paragraph quotes someone or not. Specifically, we identify quotation marks in news paragraphs and use $\mathcal{R}6$ to connect $\mathcal{V}1$ with $\mathcal{V}6$ based on whether direct quotation is detected.

\subsubsection{Entity Cues}
\noindent \underline{\textit{$\mathcal{V}6$ and $\mathcal{R}6$: Entity Cues}} We follow previous works \citep{feng2021knowledge,hu2021knowledgefakenews} to use one node to represent each entity in the external knowledge graph. We adopt TransE \citep{TransE} to learn knowledge graph embeddings and use them as initial node features for $\mathcal{V}6$. We then adopt entity linking tool TagMe \cite{tagme} to align news paragraphs with their mentioned entities and use $\mathcal{R}6$ to connect $\mathcal{V}1$ with $\mathcal{V}6$ correspondingly.

In this way, we obtain a heterogeneous information network for news articles that jointly models knowledge-enriched news content and diversified textual cues in news articles. Our approach could be similarly extended to other textual cues and paragraph-level labels that would be helpful in political perspective detection and related tasks.

\subsection{Learning and Optimization}
Upon obtaining the news HINs, we adopt relational graph neural networks for representation learning and conduct political perspective detection as graph-level classification. Specifically, we follow \citet{feng2021knowledge} and use gated R-GCN to ensure a fair comparison and highlight the effectiveness of knowledge walks and textual cues.
After $L$ layers of gated R-GCN, we denote the learned node representations as $\overline{v}$ and obtain graph-level representation $v_g$ 
with three different aggregation strategies: Paragraph Average (PA), Cue Average (CA) and Global Average (GA):
\begin{equation}
  v_g =
    \begin{cases}
      \frac{1}{|\mathcal{V}1|} \sum_{v \in \mathcal{V}1} \overline{v} & \text{if Paragraph Average;}\\
      \frac{1}{|\mathcal{V}-\mathcal{V}1|} \sum_{v \notin \mathcal{V}1} \overline{v} & \text{if Cue Average;}\\
      \frac{1}{|\mathcal{V}|} \sum_{v \in\mathcal{V}} \overline{v} & \text{if Global Average.}\\
    \end{cases}       
\end{equation}
where $\mathcal{V} = \bigcup_{i=1}^6 \mathcal{V}i$ represents the set of all nodes in our HIN. We then transform the graph-level representation $v_g$ with a softmax layer and classify news articles into perspective labels:

\begin{equation}
    \hat{y} = softmax(W_o \cdot v_g + b_o)
\end{equation}
where $W_o$ and $b_o$ are learnable parameters and $\hat{y}$ is our model's prediction. We optimize the end-to-end process with cross entropy loss and $L_2$ regularization.


\section{Experiments}

\subsection{Dataset}
We make use of two real-world political perspective detection datasets SemEval \citep{SemEval} and Allsides \citep{li2019encoding}, which are widely adopted in various previous works \citep{li2019encoding, li2021mean, feng2021knowledge}. We follow the same evaluation settings as in previous works so that our results are directly comparable. Section B in the appendix provides more dataset details to facilitate reproduction.

\begin{table}[]
    \centering
    \begin{tabular}{l c}
         \toprule[1.5pt] \textbf{Hyperparameter} & \textbf{Value} \\ \midrule[0.75pt]
         GNN input size & 768 \\
         GNN hidden size & 512 \\
         GNN layer $L$ & 2\\
         \# epoch & 150 \\
         batch size & 16 \\
         dropout & 0.6 \\
         \# knowledge walk & 30,114 \\
         $p(r)$ in Equ. (3) & constant $c$ \\
         \# head in Equ. (8) & SE: 8, AS: 32 \\
         $\lambda$ in Equ. (11) & 1e-4 \\
         learning rate & 1e-3 \\
         lr\_scheduler\_patience & 20 \\
         lr\_scheduler\_step & 0.1 \\
         \# early stop epoch & 40 \\
         Optimizer & Adam \\
         \bottomrule[1.5pt]
    \end{tabular}
    \caption{Hyperparameter settings of KCD. SE and AS denote the datasets SemEval and Allsides.}
    \label{tab:hyperparameter}
\end{table}

\subsection{Baselines}
We compare KCD with the following competitive baselines and state-of-the-art methods:
\begin{itemize}[leftmargin=*]
    \item \textbf{CNN} \citep{CNNglove} is the first-place solution in the SemEval 2019 Task 4 contest \citep{SemEval}. It combines convolutional neural networks with Glove \citep{CNNglove} and ELMo \citep{ELMo} for political perspective detection on the SemEval dataset.
    \item \textbf{HLSTM} \citep{HLSTM} is short for hierarchical long short-term memory networks. \citet{li2019encoding} uses HLSTMs and different word embeddings for news bias detection.
    \item \textbf{HLSTM\_Embed} and \textbf{HLSTM\_Output} \citep{li2021mean} leverage entity information with masked entity models in addition to news content for political perspective detection.
    \item \textbf{Word2Vec} \citep{word2vec}, \textbf{GloVe} \citep{pennington2014glove}, \textbf{ELMo} \citep{ELMo}, pre-trained \textbf{BERT} \citep{devlin2018bert} and \textbf{RoBERTa} \citep{liu2019roberta} are leveraged by \citet{feng2021knowledge} as textual features and political perspective detection is further conducted with two fully connected layers. 
    \item \textbf{MAN} \citep{li2021mean} incorporates social and linguistic information with pre-training tasks and conducts fine-tuning on the task of political perspective detection.
    \item \textbf{KGAP} \citep{feng2021knowledge}, short for \textbf{K}nowledge \textbf{G}raph \textbf{A}ugmented \textbf{P}olitical perspective detection, leverages knowledge graphs and graph neural networks for a knowledge-aware approach. We compare our gated R-GCN based approach with KGAP's gated R-GCN setting.
\end{itemize}

\begin{table}[t]
    \centering
    \resizebox{\linewidth}{!}{
        \begin{tabular}{l l|c c|c c}
             \toprule[1.5pt] \multirow{2}{*}{\textbf{Method}} & \multirow{2}{*}{\textbf{Setting}} & \multicolumn{2}{c|}{\textbf{SemEval}} & \multicolumn{2}{c}{\textbf{AllSides}} \\ 
             & & \textbf{Acc} & \textbf{MaF} & \textbf{Acc} & \textbf{MaF} \\ \midrule[0.75pt]
             \multirow{2}{*}{\textbf{CNN}}&GloVe & $79.63$ & $N/A$ & $N/A$ & $N/A$ \\
             &ELMo & $84.04$ & $N/A$  & $N/A$  & $N/A$  \\ \midrule[0.75pt]
             \multirow{4}{*}{\textbf{HLSTM}}&GloVe & $81.58$ & $N/A$  & $N/A$  & $N/A$  \\
             &ELMo & $83.28$ & $N/A$  & $N/A$  & $N/A$  \\
             &Embed & $81.71$ & $N/A$  & $76.45$ & $74.95$ \\
             &Output & $81.25$ & $N/A$  & $76.66$ & $75.39$ \\ \midrule[0.75pt]
             \multirow{5}{*}{\textbf{Text Model}} & Word2Vec & $70.27$ & $39.37$ & $48.58$ & $34.33$ \\
             & GloVe & $80.71$ & $63.64$ & $71.01$ & $69.81$ \\
             & ELMo & $86.78$ & $80.46$ & $81.97$ & $81.15$ \\
             & BERT & $86.92$ & $80.71$ & $82.46$ & $81.77$ \\
             & RoBERTa & $87.08$ & $81.34$ & $85.35$ & $84.85$ \\
             \midrule[0.75pt]
             \multirow{3}{*}{\textbf{MAN}}&GloVe & $81.58$ & $79.29$ & $78.29$ & $76.96$ \\
             &ELMo & $84.66$ & $83.09$ & $81.41$ & $80.44$ \\
             &Ensemble & $86.21$ & $84.33$ & $85.00$ & $84.25$ \\ \midrule[0.75pt]
             \textbf{KGAP}&GRGCN & $89.56$ & $84.94$ & $86.02$ & $85.52$ \\ \midrule[0.75pt]
             \multirow{3}{*}{\textbf{KCD}} 
             & GA & $88.52$ & $84.13$ & $86.02$ & $85.53$ \\
             & CA & $89.77$ & $85.26$ & $81.28$ & $80.39$ \\
             & PA & $\textbf{90.87}$ & $\textbf{87.87}$ & $\textbf{87.38}$ & $\textbf{87.14}$ \\ \midrule[0.75pt]
             \multirow{2}{*}{\textbf{KCD (PA)}} 
             & - w/o TC & $88.22$ & $83.53$ & $86.08$ & $85.58$ \\
             & - w/o KW & $87.29$ & $81.77$ & $85.51$ & $85.00$ \\
             \bottomrule[1.5pt]
        \end{tabular}
    }
    \caption{Political perspective detection performance on two benchmark datasets. Acc and MaF denote accuracy and macro-averaged F1-score. N/A indicates that the result is not reported in previous works. TC and KW indicate textual cues and knowledge walks respectively.}
    \label{tab:big}
\end{table}

\subsection{Implementation}
We implement our KCD framework with pytorch \citep{NEURIPS2019_bdbca288}, pytorch lightning \citep{Falcon_PyTorch_Lightning_2019}, pytorch geometric \citep{torchgeometric}, and the transformers library \citep{wolf-etal-2020-transformers}. We present our hyperparameter settings in Table \ref{tab:hyperparameter} to facilitate reproduction. We adhere to these settings throughout all experiments in the paper unless stated otherwise. Our implementation is trained on a Titan X GPU with 12GB memory. We make our code and data publicly available\footnote{https://github.com/Wenqian-Zhang/KCD}.

\subsection{Experiment Results}
We present model performance on two benchmark datasets in Table \ref{tab:big}, which demonstrates that
\begin{itemize}[leftmargin=*]
    \item KCD, especially with the PA aggregation strategy, consistently outperforms state-of-the-art methods on both benchmark datasets.
    \item KGAP and KCD, which incorporate knowledge graphs, outperform other baselines. This indicates that external knowledge is essential in providing background information and political context to analyze ideological perspectives. KCD without textual cues performs better than baseline methods except KGAP and performs close to KGAP. These suggests that KGAP's method of infusing knowledge as HIN nodes and our method of infusing knowledge as knowledge walks are both effective.
    \item PA outperforms CA and GA on both datasets, which suggests that the aggregation strategy is important since subsidiary nodes like textual cues may result in noise. As a result, we should focus on paragraph nodes in our heterogeneous information networks. 
    \item Removing textual cues and knowledge walks in KCD result in substantial performance drop, which demonstrates the effectiveness of textual cues and knowledge walks. 
\end{itemize}

\begin{figure}
    \centering
    \includegraphics[width=0.9\linewidth]{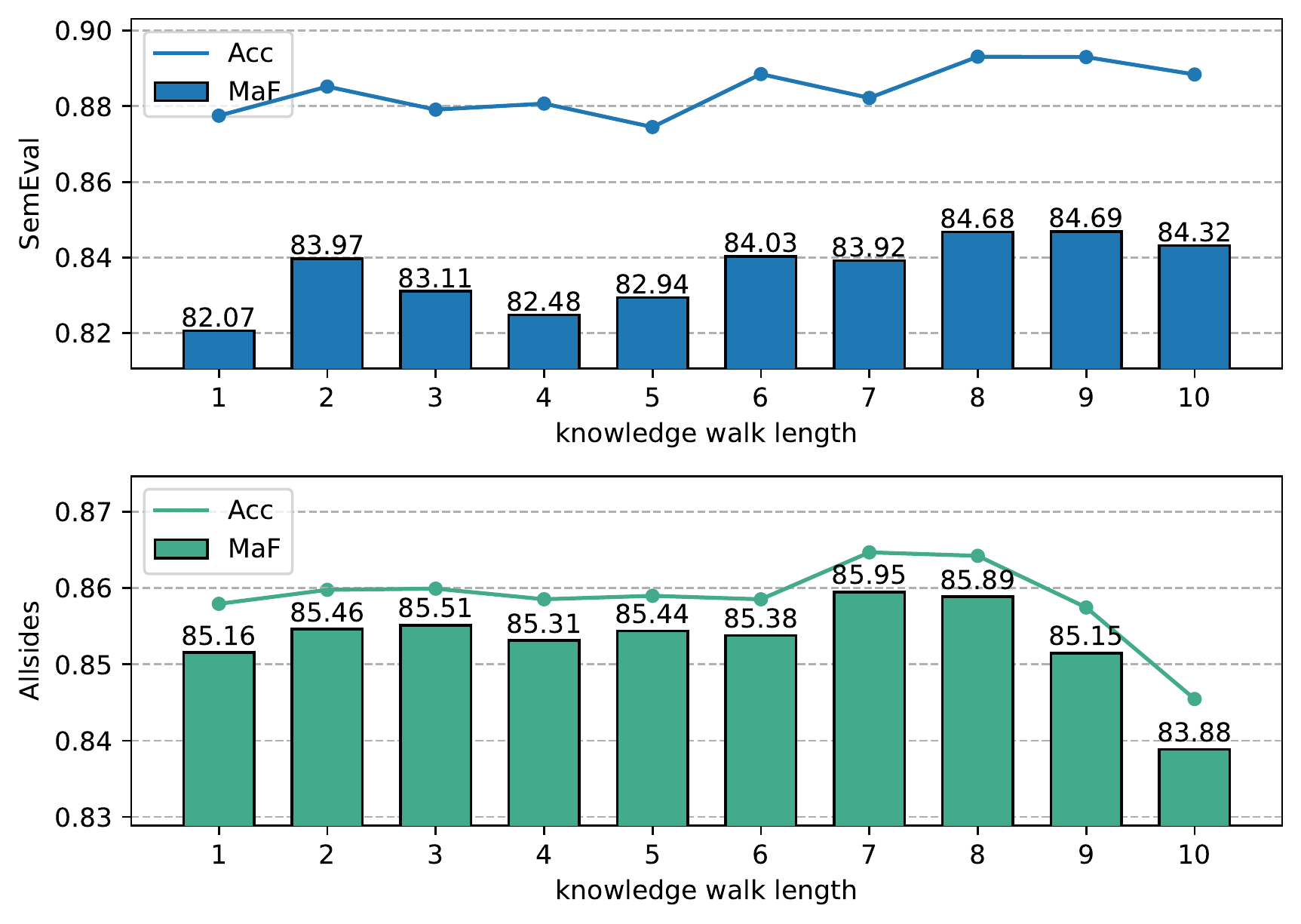}
    \caption{Our approach's performance when the maximum length of knowledge walk generation is specified from 1 to 10 knowledge graph triples.}
    \label{fig:knowledge_walk_length}
\end{figure}

In the following, we examine the effect of knowledge walks and textual cues in our approach. We also explore how our approach performs with limited data compared to baseline methods.

\subsection{Knowledge Walks Study}
We propose knowledge walks, an approach to conduct multi-hop reasoning on knowledge graphs and inject them into textual representations. We study the effect of knowledge walk length and knowledge infusion strategies on our model's performance.


\subsubsection{Knowledge Walks Length}
Our proposed knowledge walks could be of any length, where shorter walks provide more condensed knowledge and longer walks provide more diverse knowledge. To examine the effect of knowledge walk length, we generate 5,088\footnote{so that there is a knowledge walk beginning with every possible (entity, relation) in the knowledge graph.} knowledge walks of 1 to 10 triples and present model performance in Figure \ref{fig:knowledge_walk_length}. It is illustrated that longer knowledge walks (8 or 9 for SemEval, 7 or 8 for Allsides) perform better than shorter ones, indicating the necessity of multi-hop knowledge reasoning in the task of political perspective detection.

\subsubsection{Knowledge Infusion Strategy}
We propose a two-step approach to infuse multi-hop knowledge reasoning into textual representations of news articles:
\begin{itemize}[leftmargin=*]
    \item First Aggregation: We firstly aggregate different generated knowledge walks based on semantic relevance in Equ. (\ref{equ:first_agg_start}) and Equ. (\ref{equ:first_agg_end}).
    \item Second Aggregation: We then use multi-head attention to aggregate all paragraphs and knowledge representations with Equ. (\ref{equ:second_agg_start}) and Equ. (\ref{equ:second_agg_end}).
\end{itemize}

\begin{figure}[t]
    \centering
    \includegraphics[width=1\linewidth]{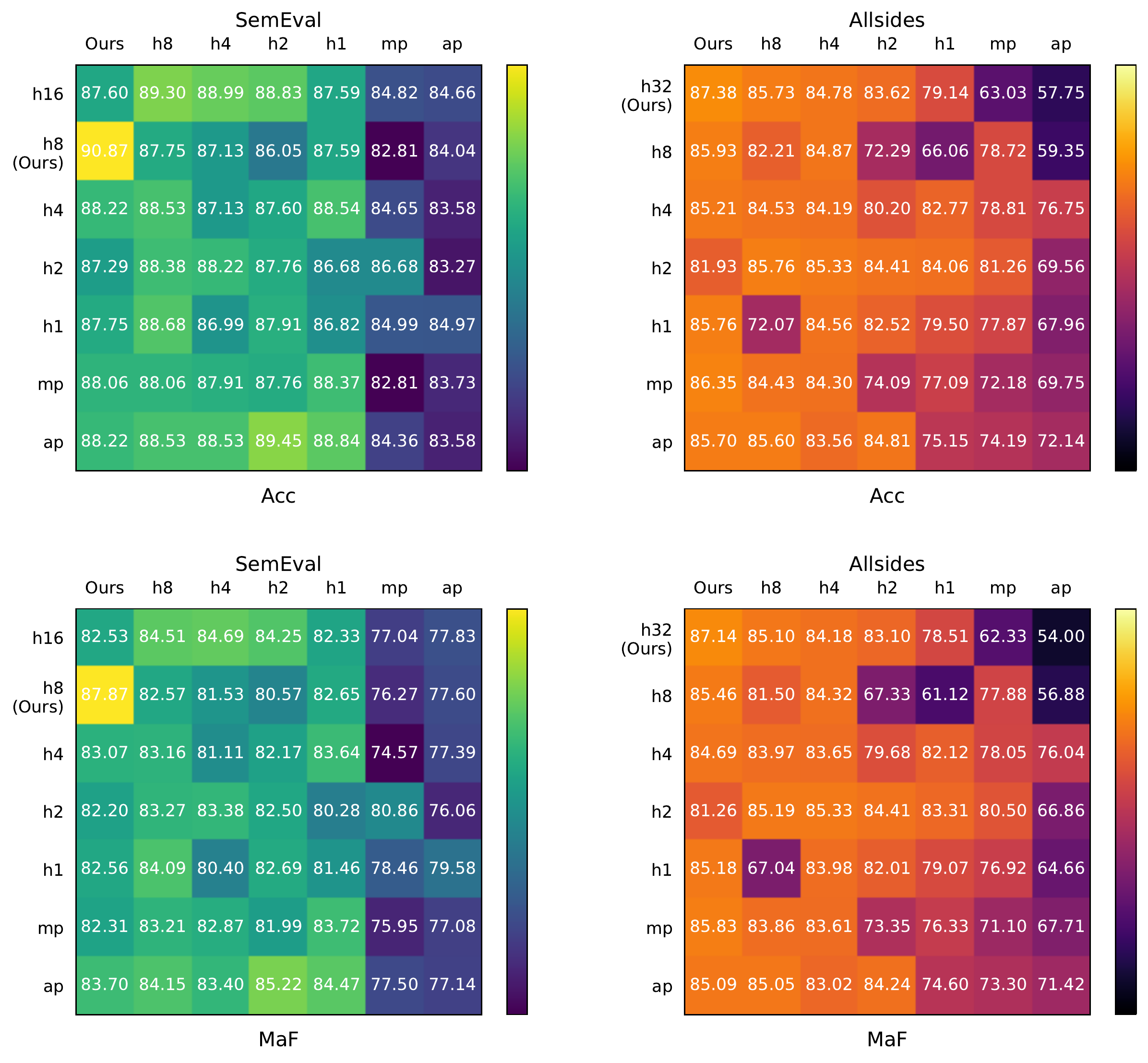}
    \caption{Model performance with different knowledge infusion strategies at two aggregation steps. The horizontal and vertical axis represent the first and second aggregation. $h_k$ denotes multi-head attention with $k$ heads, mp and ap stand for max and average pooling.}
    \label{fig:knowledge_walk_infusion}
\end{figure}

\begin{figure*}
    \centering
    \includegraphics[width=0.9\linewidth]{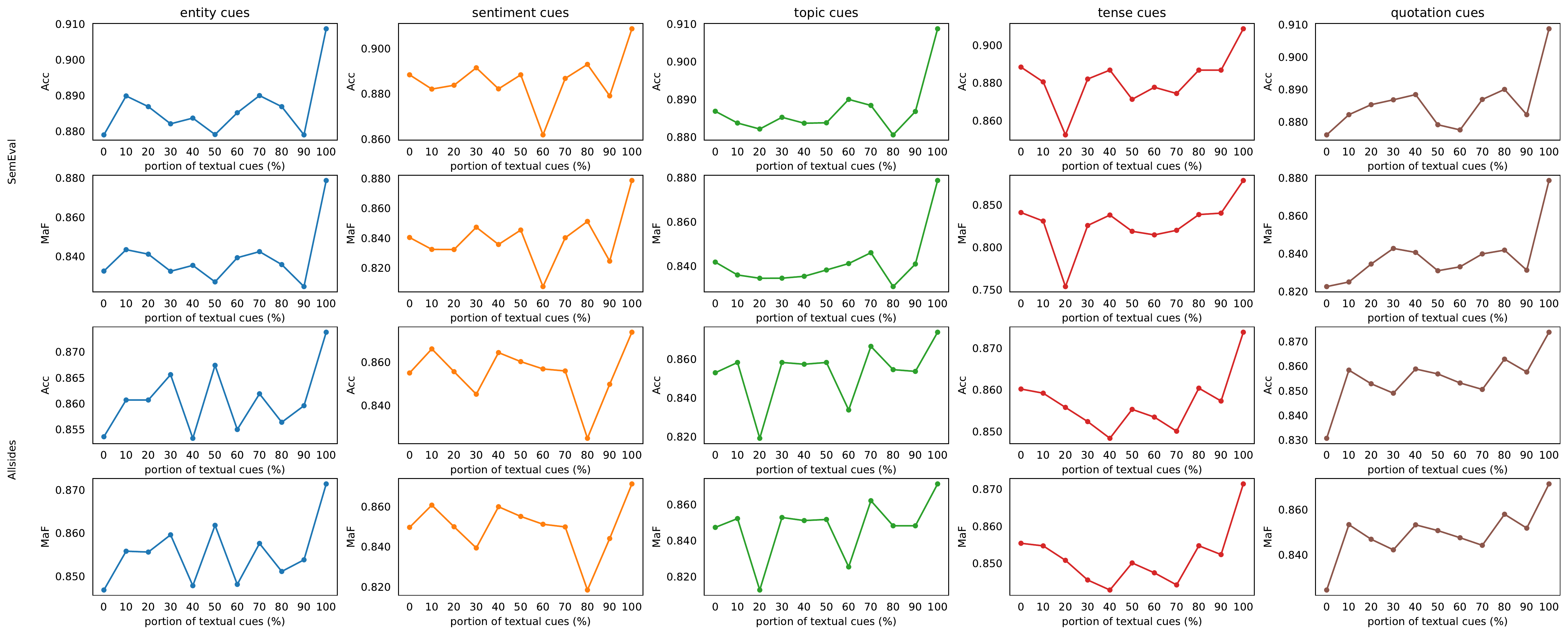}
    \caption{Model performance when five different types of textual cues are gradually removed.}
    \label{fig:textual_cue_remove}
\end{figure*}

To examine the effect of our knowledge infusion strategy, we substitute these two aggregation steps with different multi-head attention settings as well as max and average pooling. 
Results in Figure \ref{fig:knowledge_walk_infusion} demonstrate significant performance difference on the horizontal axis. This suggests that our semantic relevance-based knowledge walks aggregation strategy in Equ. (\ref{equ:first_agg_start}) and Equ. (\ref{equ:first_agg_end}) successfully filters out irrelevant knowledge reasoning and contributes to model performance. Besides, according to the vertical axis, our adopted multi-head attention in Equ. (\ref{equ:second_agg_start}) and Equ. (\ref{equ:second_agg_end}) is generally effective and does not rely on specific attention head settings.

\subsection{Textual Cues Study}
We propose to leverage semantic, syntactic and entity textual cues as paragraph-level labels to leverage implicit indicators in news articles for political perspective detection. To examine the effectiveness of these textual cues, we randomly remove them with probability $p$ and present model performance in Figure \ref{fig:textual_cue_remove}. It is illustrated that:
\begin{itemize}[leftmargin=*]
    \item A performance boost is observed between $0\%$ and $100\%$ for all five textual cues, suggesting the necessity of modeling implicit textual indicators. Besides, adding only part of textual cues sometimes leads to a decrease in performance, which implies that incomplete cues may be counter-productive and introduce noise.
    \item Among five different cues, entity and quotation cues contribute more to model performance than others. This suggests some implicit textual cues are more important than others in analyzing the ideological perspectives of news articles.
    \item The effect of textual cues is larger on the dataset SemEval, which is significantly smaller than Allsides. This suggests that we alleviate the data-hungry problem by introducing diversified textual cues as paragraph-level labels and contribute to model performance.
    
\end{itemize}

\subsection{Data Efficiency Study}
\begin{figure}
    \centering
    \includegraphics[width=1\linewidth]{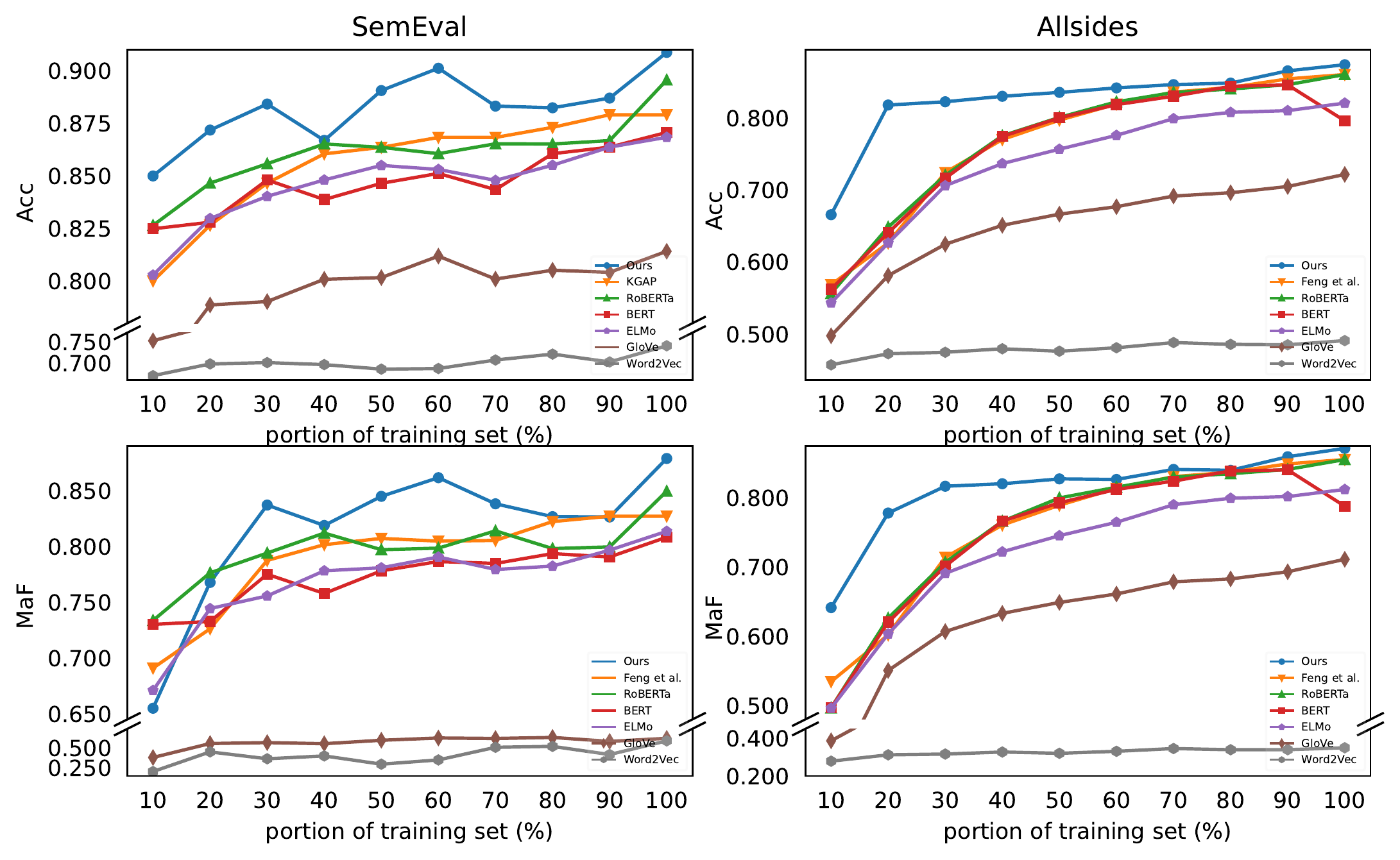}
    \caption{Model performance when KCD and various competitive baselines are trained with $10\%$ to $100\%$ of the training set on SemEval and Allsides.}
    \label{fig:data_efficiency}
\end{figure}

As \citet{li2021mean} point out, supervised data annotations could be difficult and expensive to obtain for the task of political perspective detection in news media. Our proposed knowledge walks and textual cues serve as additional information and might help mitigate this issue. To examine whether we have achieved this end, we train KCD, KGAP \citep{feng2021knowledge} as well as various text models with reduced training sets of SemEval and Allsides. Results in Figure \ref{fig:data_efficiency} demonstrate that
\begin{itemize}[leftmargin=*]
    \item KCD has better data efficiency and achieves steady performance with smaller training sets. This observation is especially salient on Allsides where the news articles are longer \citep{li2021mean}, thus more knowledge walks and textual cues could be extracted and incorporated to alleviate data dependence.
    \item Both KCD and KGAP leverage external knowledge and are more robust to reduced datasets. Our approach further leverages textual cues and has better data deficiency. This suggests that a solution to limited data could be incorporating information in addition to news content.
    \item With only 10\% training set, KCD outperforms all baselines by at least 5.68\% and 9.71\% in accuracy on two datasets. This suggests that our approach is simple, effective, and not data-hungry under limited data settings.
\end{itemize}

\section{Conclusion}
In this paper,we propose KCD, a political perspective detection approach that reasons with multi-hop external knowledge and leverages diversified implicit textual indicators. We firstly generate multi-hop knowledge walks, dynamically aggregate them based on semantic relevance and infuse into news text representations. We then construct a heterogeneous information network to jointly model knowledge-enriched news content and diversified textual cues as paragraph-level labels. Finally, we learn graph representations with relational graph neural networks under different aggregation settings and conduct political perspective detection as graph-level classification. Extensive experiments demonstrate that our approach consistently outperforms state-of-the-art baselines on two benchmark datasets. Further experiments also bear out the necessity of knowledge walks and textual cues in modeling political perspectives in news media.

\section{Acknowledgment}
We would like to thank the reviewers and chairs for their feedback and suggestions. We would also like to thank all LUD Lab members for our collaborative research environment. This study was supported by the National Key Research and Development Program of China (No. 2020AAA0108800), National Nature Science Foundation of China (No. 61872287, No. 62192781, No. 61937001, No. 62050194, No. 62137002), Innovative Research Group of the National Natural Science Foundation of China (61721002), Innovation Research Team of Ministry of Education (IRT\_17R86), Project of China Knowledge Center for Engineering Science and Technology, and Project of Chinese academy of engineering ``The Online and Offline Mixed Educational Service System for ‘The Belt and Road’ Training in MOOC China''.


\bibliography{custom}
\bibliographystyle{acl_natbib}


\appendix
\section{Limitations}
Our proposed model has two minor limitations:
\begin{itemize}[leftmargin=*]
    \item We propose to model news articles with heterogeneous information networks. This graph-based approach might not fit well with shorter news articles with only a few paragraphs. This issue might be addressed by using sentence nodes instead of paragraph nodes for shorter articles.
    \item For very large knowledge graphs with many different types of relations, it might be hard for domain experts to help set $p(r)$ for every knowledge graph relation. This issue might be addressed by only setting a larger $p(r)$ for several important $r$s according to domain expert.
\end{itemize}

\section{Dataset Details}
We used the same datasets as in previous works \citep{li2019encoding,li2021mean,feng2021knowledge}, namely SemEval \citep{SemEval} and Allsides \citep{li2019encoding}. We follow the same 10-fold setting for SemEval and 3-fold setting for Allsides \citep{li2021mean}. We use the exact same folds so that the results are directly comparable. A minor difference would be that we have to discard a few news articles on Allsides since their urls have expired and we could not retrieve their original news article. We report the statistical information of SemEval and Allsides in Table \ref{tab:dataset_detail}.

\section{Computation Details}

\subsection{Computational Resources}
Our proposed approach has a total of 7.8M learnable parameters. It takes approximately 0.7 and 1.6 GPU hours to train our approach on two datasets respectively. We train our model on one Titan X GPU with 12GB memory.

\subsection{Experiment Runs}
We run our approach with three different aggregation strategies \textbf{five times} and report the average accuracy and macro F1-score with standard deviation in Table \ref{tab:five_run}. For experiments in Section 4.5, 4.6 and 4.7, we do not have enough computational resources to run five times, thus we report the performance of a single run.

\begin{table}[]
    \centering
    \resizebox{\linewidth}{!}{
    \begin{tabular}{l c c c}
         \toprule[1.5pt] \textbf{Dataset} & \textbf{\# Articles} & \textbf{\# Class} & \textbf{Class Distribution} \\ \midrule[0.75pt]
         SemEval & 645 & 2 & 407 / 238  \\
         Allsides & 10,385 & 3 & 4,164 / 3,931 / 2,290  \\ \bottomrule[1.5pt]
    \end{tabular}
    }
    \caption{Details of two datasets SemEval and Allsides.}
    \label{tab:dataset_detail}
\end{table}

\begin{table}[]
    \centering
    \resizebox{\linewidth}{!}{
        \begin{tabular}{l c|c c}
            \toprule[1.5pt] 
            \textbf{Datasets} & \textbf{Setting}
            & \textbf{Acc} & \textbf{MaF}  \\ 
             
            \midrule[0.75pt]
            \multirow{3}{*}{\textbf{SemEval}} 
             & GA & $88.62 \pm 0.32$ & $83.86 \pm 0.27$ \\
             & CA & $89.30 \pm 0.24$ & $84.85 \pm 0.30$ \\
             & PA & $89.90 \pm 0.68$ & $86.11 \pm 1.18$ \\
            \midrule[0.75pt]
            \multirow{3}{*}{\textbf{AllSides}} 
             & GA & $84.88 \pm 2.90$ & $84.31 \pm 2.96$ \\
             & CA & $79.85 \pm 2.58$ & $79.41 \pm 2.56$ \\
             & PA & $87.17 \pm 0.24$ & $86.72 \pm 0.35$ \\ 
             \bottomrule[1.5pt]
        \end{tabular}
    }
    \caption{Average performance and standard deviation of three different aggregation strategies for five runs.}
    \label{tab:five_run}
\end{table}
\section{Scientific Artifact Usage}
We provide additional details about used scientific artifacts and specifically how we used them.

\balance

\begin{itemize}[leftmargin=*]
    \item NLTK \citep{nltk}: We use NLTK to extract the tense of news articles. Specifically, we first use NLTK POS-tagger to process new paragraphs and attach speech tag to each word. Then we align verb tags with NLTK tagset to identify the tense of paragraphs. 
    \item BertTopic \citep{grootendorst2020bertopic}: We use BertTopic to mine the topics of news corpus. Specifically, we use BertTopic topic model to learn dataset-specific topic models. For SemEval we obtained 197 topics and for Allsides we obtained 1225 topics. Next, we predict topics for each news paragraph. Each topic consists of ten topic words with scores and we select the top five to serve as the news paragraph's topic.
    \item Huggingface Transformers \citep{wolf-etal-2020-transformers}: We use the pipeline module for sentiment analysis. Specifically, we use the sentiment analysis API in the text classification pipeline to generate a sentiment label and score for news paragraphs. We then use the sentiment label as the sentiment cues for news paragraphs.
    \item TagMe \citep{tagme}: We use TagMe to align news articles with entities in the knowledge graph. Specifically, we use TagMe to annotate named entities in news paragraphs and save the entities with a score higher than 0.1 for further alignment. We then calculate the similarity score between TagMe annotated entities and political knowledge graph entities. We recognize the entities with a score higher than 0.9 as entity cues in our constructed HIN.
    \item Political knowledge graph \citep{feng2021knowledge}: We use the political knowledge graph collected in \citet{feng2021knowledge} for external knowledge in political perspective detection.
    \item OpenKE \citep{han-etal-2018-openke}: We use OpenKE to train TransE \citep{TransE} knowledge graph embeddings for the political knowledge graph. Specifically, we set the TransE hidden size to 768 and train the model with other default hyperparameters in OpenKE.
\end{itemize}

\end{document}